\documentclass[]{article}
\usepackage[affil-it]{authblk}
\usepackage{graphicx}
\usepackage{booktabs}
\usepackage[hyphens]{url}
\usepackage{hyperref}

\providecommand{\keywords}[1]
{
  \small	
  \textbf{\textit{Keywords---}} #1
}

\title{Comparison of Czech Transformers on Text Classification Tasks}
\author{Jan Lehe\v{c}ka$^{1}$, Jan \v{S}vec$^{1}$  \\
        \small $^{1}$Department of Cybernetics, University of West Bohemia in Pilsen, Univerzitn\'{i} 2732/8, 301 00 Pilsen, Czech Republic\\
$\{$jlehecka,honzas$\}$@kky.zcu.cz
}
\date{} 

\begin{document}

\maketitle

\begin{abstract}
In this paper, we present our progress in pre-training monolingual Transformers for Czech and contribute to the research community by releasing our models for public. The need for such models emerged from our effort to employ Transformers in our language-specific tasks, but we found the performance of the published multilingual models to be very limited. Since the multilingual models are usually pre-trained from 100+ languages, most of low-resourced languages (including Czech) are under-represented in these models. At the same time, there is a huge amount of monolingual training data available in web archives like Common Crawl. We have pre-trained and publicly released two monolingual Czech Transformers and compared them with relevant public models, trained (at least partially) for Czech. The paper presents the Transformers pre-training procedure as well as a comparison of pre-trained models on text classification task from various domains.

\end{abstract}

\section{Introduction}
In the last few years, deep neural networks based on Transformers~\cite{devlin-etal-2019-bert} have dominated the research field of Natural Language Processing (NLP) and Natural Language Understanding (NLU). Self-attention~\cite{vaswani2017attention} Transformers and especially the self-supervised-trained variants known as BERT (Bidirectional Encoder Representations from Transformers) models, have achieved amazing results in many tasks, including text classification~\cite{sun2019fine,adhikari2019docbert,lehevcka2020adjusting}.

Because the self-supervised pre-training of Transformers is computationally very costly, researchers around the world publish their pre-trained models in order to compete on a large variety of NLP and NLU tasks, however the majority of these models are monolingual English models or multilingual models including many languages at once. Since the multilingual models are usually pre-trained from Wikipedia dumps or large web archives, most of low-resourced languages are under-represented in these models. This makes practical use of such models in non-English languages very limited.
Therefore in present days, we observe a significant trend in pre-training also monolingual transformers for other languages, e.g. CamemBERT for French \cite{martin2019camembert}, Finnish BERT \cite{virtanen2019multilingual} and very recently also Czech models Czert \cite{sido2021czert} and RobeCzech \cite{straka2021robeczech}. 

The trend in pre-training monolingual Transformers is boosted by the accessibility of powerful hardware devices (especially graphical and text processing units, GPUs and TPUs) and by the availability of huge text corpora even for languages with relatively small number of native speakers. For example, many researchers use data from the Common Crawl project\footnote{\url{https://commoncrawl.org}} which is a huge public web archive consisting of petabytes of crawled web pages.

In this paper, we present our effort to pre-train two monolingual Czech Transformers (BERT and RoBERTa) from different datasets and we compare the performance with published multilingual and monolingual Czech models on downstream text classification  tasks. We release both our pre-trained models publicly on HuggingFace hub\footnote{\url{https://huggingface.co/fav-kky}}.

\section{Related Work}

Aside from the famous English BERT-base model \cite{devlin-etal-2019-bert}, also its multilingual variant has been published by Google researchers. It has been trained from Wikipedia dumps of $104$~languages, including Czech.
We denote this model \textit{MultiBERT} in this paper.

BERT model trained from Wikipedia dumps of only four Slavic languages (Russian, Bulgarian, Czech and  Polish) has been presented in \cite{arkhipov2019tuning}. The model was not pre-trained from the scratch, but MultiBERT weights were used to initialize the model. We denote this model \textit{SlavicBERT} in this paper.

Researchers from Facebook have published multilingual \textit{XLM-RoBERTa} model \cite{conneau2020unsupervised} pre-trained on one hundred languages (including Czech), using more than two terabytes of filtered Common Crawl data. Two sizes of the model have been published: base model with 270 million parameters and a large one with 550 million parameters.

There are also two recent papers presenting monolingual Czech models. \textit{Czert} \cite{sido2021czert} is a BERT model pre-trained from a mixture of Czech national corpus \cite{11234/1-1846}, Wikipedia pages and self-crawled news dataset.
The second one, \textit{RobeCzech} \cite{straka2021robeczech}, is a RoBERTa \cite{liu2019roberta} model pre-trained from Czech national corpus \cite{11234/1-1846}, Wikipedia pages, a collection of Czech newspaper and magazine articles and a Czech part of the web corpus W2C \cite{11858/00-097C-0000-0022-6133-9}.
We provide the basic information about described Transformers in the Tab.~\ref{tab:models}.

\section{Datasets}

\subsection{Pre-training Datasets}
Since self-supervised pre-training of Transformers requires a huge amount of unlabeled text, our aim was to get as much cleaned Czech text as possible. In the following paragraphs, we provide a description of datasets we were experimenting with.

\subsubsection{News Corpus}
We used Czech news dataset from our web-mining framework \cite{Svec2014a}. The dataset consists mainly of web-crawled news articles and transcripts of selected TV and radio shows. In this dataset, we cleaned boilerplate and HTML tags from original documents and thoroughly processed each document with normalization, true-casing and rule-based word substitution dealing with typos, multiwords and word-form unification. See \cite{Svec2014a} for more details. The corpus contains 3.3 billion words (20.5~GB of cleaned text).

\subsubsection{C5}
Czech Colossal Clean Crawled Corpus (C5) generated from the Common Crawl project is a Czech mutation of the English C4 dataset \cite{2019t5}. We used the language information provided in the index files to select Czech records only. The corresponding plain texts (stored in WET archives) were downloaded. To clean the data, we followed almost the same rules which were used to pre-process the C4 dataset, i.e.:
\begin{itemize}
    \item We only retained lines that ended with a terminal punctuation mark (".", "?" or "!").
    \item We removed lines containing "javascript" or "cookies".
    \item We removed web pages containing offensive words or strings "lorem ipsum" or "\{".
    \item We retained only pages classified as Czech with probability of at least 0.99 according to \texttt{langdetect}\footnote{\url{https://pypi.org/project/langdetect/}} tool.
    \item To deduplicate the dataset, we discarded all but one line occurring more than once in the data set.
    \item We only retained lines with at least 3 words and pages with at least 5 sentences.
\end{itemize}
This simple yet rigorous cleaning process removed about 98\% of downloaded plain texts from the dataset, mainly due to the deduplication (the more data we downloaded, the harder it was to find a new unobserved line).

We downloaded and cleaned recent crawls up to the crawl from August 2018, which is the first one providing the language column in its index. Together, we processed 25 crawls (from August 2018 to October 2020) and the resulted dataset contains almost 13 billion words (93~GB of cleaned text\footnote{To get a picture how huge the Common Crawl archive is, consider that this 93~GB of text is (due to rigorous cleaning rules) just about 2\% of all Czech texts we downloaded and Czech records occupy just about 1\% of the full archive.}).

\subsection{Text Classification Datasets}
We decided to fine-tune and evaluate Transformers on a text classification task. Specifically, we fine-tuned models for sentiment analysis and multi-label topic identification tasks as these tasks are in our main focus. To keep our experiments reproducible, we experimented mainly with publicly available datasets. In the following paragraphs, we describe five Czech datasets used in the experiments together with the basic statistics of the datasets in Tab.~\ref{tab:evalDatasets}. The first three datasets are for sentiment analysis task. All of them were created in~\cite{habernal2013sentiment} and contain text samples with positive, neutral or negative class labels. The last two datasets are for multi-label topic identification task. Both of them are from news domain and contain crawled news articles, each with at least one topic label.

\begin{table}[th]
  \centering
  \begin{tabular}{ l r r r r r}
    \toprule
    \textbf{Dataset} & \textbf{Train} & \textbf{Devel} & \textbf{Test} & \textbf{Classes} & \textbf{LabCard}\\
    \midrule
    CSFD &  $91\,381$ & - & - & $3$ & $1$ \\
    MALL & $145\,307$ & - & - & $3$ & $1$ \\
    FCB  &   $9\,752$ & - & - & $3$ & $1$ \\
    \midrule
    CN   & $184\,313$ & $20\,479$ & $43\,762$ & $577$ & $3.06$ \\
    CTDC & $11\,955$ & $2\,538$ & - & $37$ & $2.55$ \\
    \bottomrule
  \end{tabular}
  \caption{Czech text classification datasets statistics. The table shows numbers of documents in train, development and test datasets, number of classes and the label cardinality (i.e. the average number of labels per document) of the datasets.}
  \label{tab:evalDatasets}
\end{table}

\subsubsection{CSFD}
Movie review dataset consists of 91 thousand movie reviews from the Czech-Slovak Movie Database (\v{C}SFD)\footnote{\url{https://www.csfd.cz}}. Based on movie ratings, each review was classified into positive, neutral or negative sentiment class.

\subsubsection{MALL}
Product review dataset consists of 145 thousand posts from a large Czech e-shop Mall.cz\footnote{\url{https://www.mall.cz}} which offers a wide range of products. Similar to CSFD dataset, each post was classified into positive, neutral or negative sentiment class based on the user's ratings.

\subsubsection{FCB}
The Facebook dataset contains 10 thousand posts selected from several Facebook pages with a large Czech fan base. The sentiment of individual posts was assigned manually by multiple annotators. We ignored the 248 bipolar posts and trained models only with three sentiment classes as suggested by authors of the dataset \cite{habernal2013sentiment}.

\subsubsection{CTDC}
Czech Text Document Corpus (CTDC)\footnote{ \url{http://ctdc.kiv.zcu.cz}} is a news dataset described in \cite{KRAL18.671}. The dataset consists of $12$~thousand news articles annotated with category labels. In order to keep our results comparable with other papers (such as \cite{lenc2016deep,sido2021czert}), we used the same setup and selected only 37 (out of 60) most frequent labels.

\subsubsection{CN}
Many news servers publish articles along with topic labels assigned by journalists. We keep this information in our framework \cite{Svec2014a} and map the most frequent labels into a standardized IPTC news topic tree\footnote{\url{http://cv.iptc.org/newscodes/mediatopic}}. We created a CN dataset by selecting articles with assigned IPTC topic labels from one news server (\url{ceskenoviny.cz}, hence the name CN) which we believe to be labeled thoroughly and consistently. The dataset contains $250$~thousand articles from 10-years epoch. Each article has assigned one or more topic labels. The total number of labels in the dataset is $577$. Unfortunately, we do not have licence to publish this dataset. It is, however, the only private dataset appearing in this paper.

\section{Models}
In this paper, we present two new Transformers. We named our models Flexible Embedding Representation NETwork (FERNET).
We tabulate basic summary of them along with all other models we have experimented with in Tab.~\ref{tab:models}, and describe them in the following paragraphs.

\begin{table*}[tb]
  \centering
  \resizebox{\textwidth}{!}{\begin{tabular}{ l c c c c c c}
    \toprule
    \textbf{Model} & \textbf{Architecture} & \textbf{Tokenization} & \textbf{Vocab} & \textbf{Pre-train data} & \textbf{Params} \\
    \midrule
    MultiBERT \cite{devlin-etal-2019-bert} & BERT-base & WordPiece & $120$~k & Wiki, 104 langs & $179$~M \\
    SlavicBERT \cite{arkhipov2019tuning} & BERT-base & subword-nmt & $120$~k & Wiki, 4 langs & $179$~M \\
    XLM-RoBERTa-base \cite{conneau2020unsupervised} & RoBERTa-base & SPM & $250$~k & CC, 100 langs (2TB) & $278$~M \\
    XLM-RoBERTa-large \cite{conneau2020unsupervised} & RoBERTa-large & SPM & $250$~k & CC, 100 langs (2TB) & $560$~M \\
    \midrule
    Czert \cite{sido2021czert} & BERT-base & WordPiece & $40$~k & Nat+Wiki+News (37GB) & $110$~M \\
    RobeCzech \cite{straka2021robeczech} & RoBERTa-base & BBPE & $52$~k & Nat+Wiki+Czes+W2C & $126$~M \\
    \midrule
    FERNET-C5 & BERT-base & SPM & $100$~k & C5 (93GB) & $164$~M \\
    FERNET-News & RoBERTa-base & BBPE & $50$~k & News Corpus (21GB) & $124$~M \\
    \bottomrule
  \end{tabular}}
  \caption{Summary of pre-trained Transformers trained at least partially on Czech. The first four models are multilingual, the next two models are already published monolingual Czech Transformers. The two models at the bottom of the table are our new models presented in this paper. The sizes of vocabularies are in thousands (k) and number of trainable parameters are in millions (M). Reported numbers of parameters include language model head on top of the model. SPM stands for SentencePiece tokenization \cite{kudo2018sentencepiece}, BBPE for Byte-level Byte-Pair Encoding and CC for Common Crawl.}
  \label{tab:models}
\end{table*}

\subsubsection{FERNET-C5}
This is a BERT model trained from scratch using the C5 dataset. The model has the same architecture as BERT-base model \cite{devlin-etal-2019-bert}, i.e. $12$ transformation blocks, $12$ attention heads and the hidden size of $768$ neurons. 

In contrast to Google's BERT models, we used SentencePiece tokenization \cite{kudo2018sentencepiece} instead of the Google's internal WordPiece tokenization. We trained a BPE SentencePiece model from the underlying dataset with vocabulary size set to 100~thousands. Since the datasets contains a small portion of non-Czech text fragments, we carefully tuned the character coverage parameter to fully cover all Czech characters and only reasonable number of graphemes from other languages. We kept original casing of the dataset.

\subsubsection{FERNET-News}
This is a RoBERTa model trained from scratch using our thoroughly pre-processed News dataset. The model has the same architecture as RoBERTa-base model \cite{liu2019roberta}, i.e. $12$ transformation blocks, $12$ attention heads and the hidden size of $768$ neurons. 
To tokenize input texts, we trained byte-level BPE tokenizer with the vocabulary size 50\,000 tokens. 

\subsection{Pre-training}
Our BERT model was pre-trained using two standard tasks: masked language modeling (MLM), which is a fill-in-the-blank task, where the model is taught to predict the masked words, and next sentence prediction (NSP) task, which is the classification task of distinguishing between inputs with two consequent sentences and inputs with two randomly chosen sentences from a corpus.
We pre-trained the model for $6.5$~million gradient steps in total ($3.5$~M steps with 128-long inputs and batch size $256$, and $3$~M steps with $512$-long inputs and batch size $128$). For both input variants, we used whole word masking, duplication factor of 5 and the learning rate warmed up over the first 100\,000~steps to a peak value at $1\times10^{-4}$.
To pre-train the model, we used the software provided by Google researchers\footnote{\url{https://github.com/google-research/bert}} and to speed up the training, we used one 8-core TPU with 128\,GB of memory. The pre-training took about two months.

Our RoBERTa model was pre-trained using MLM task only.
We trained the model for $600$~thousand gradient steps with batch size $2048$: first $500$ thousand steps with 128-long inputs, 24\,000 warmup steps and learning rate $4\times10^{-4}$, and the following $100$ thousand steps with $512$-long inputs, 2\,000 warmup steps and learning rate $1\times10^{-4}$.
Since the news dataset is rather small (but very clean), the model iterated about 40-times over the whole training corpus during the pre-training. However, due to dynamic language model masking, the model saw differently masked input samples in each epoch.
To pre-train the model, we used the HuggingFace's Transformers software\footnote{\url{https://huggingface.co/transformers}} and to speed up the training, we used two Nvidia A100 GPUs. The pre-training took about one month.

\subsection{Fine-tuning}
We run fine-tuning experiments on single GPUs using HuggingFace's Transformers software.

To fine-tune models for sentiment analysis, we fixed the peak learning rate to $1\times10^{-5}$, maximum sequence length 128 tokens, batch size 64, 10 training epochs and 100 warmup steps. We were also experimenting with other hyper-parameters setting, but this one gave consistently the best results across all models and datasets. After each training epoch, we evaluated the model on the development dataset (if not present in the dataset, we randomly held out 10\% of the training samples as a development data) and at the end, we used the best model for evaluation on the test dataset.

To fine-tune models for multi-label topic identification task, we fixed the peak learning rate to $5\times10^{-5}$, maximum sequence length 512 tokens and batch size 64. 
We trained on the CTDC dataset for 20 epochs whereas to train on CN dataset, 10 epochs was enough, because the dataset is much larger.

\subsection{Evaluation}
For both BERT and RoBERTa model architectures, we predicted classes using standard pooling layers for sequence classification implemented in the Transformers library along with the corresponding model types. This pooling layer uses the last hidden state of the first input token (\texttt{[CLS]} or \texttt{<s>}) to make predictions.

In the case of sentiment analysis task, we 
assigned sentiment class with the highest predicted value for each document and evaluated F1 score.

In the case of multi-label text classification tasks, we added a sigmoid activation function to the final output layer of each model, thus the output for each document was a vector of per-label scores $s = (s_1, s_2, ..., s_K)$, $s_i \in (0,1)$, where $K$ is total number of labels. We converted this "soft" predictions into binary "hard" predictions by following thresholding strategy: assign $i$-th label, if
\begin{equation}
    \frac{s_i}{\max(s_1, s_2, ..., s_K)} \geq 0.5.
\end{equation}
This strategy gave consistently better results than simple thresholding sigmoid outputs on $0.5$. After thresholding, we evaluated sample-averaged F1 score.

For datasets which do not contain separated test dataset for evaluation (all but CN), we performed 10-fold cross-validation. For sentiment analysis datasets, we split the data while preserving the percentage of samples for each class (i.e. stratified cross-validation), for topic identification datasets, we split the data randomly. For each pre-trained model and each dataset, we fine-tuned 10 different models (one for each fold), averaged the results and reported the mean and standard deviation.

\section{Results}
We summarize the experimental results in Tab.~\ref{tab:results_SA} (sentiment analysis task) and Tab.~\ref{tab:results_MLTI}  (multi-label topic identification task). In both tables, we first report "base-model-size" category results as these models have more or less comparable number of training parameters (see Tab.~\ref{tab:models} for details). In the last row, we report also results with XLM-RoBERTa-large model, which has, however, several-times more parameters.

\setlength{\tabcolsep}{8pt}
\begin{table}[th]
  \centering
  \begin{tabular}{ l c c c }
    \toprule
    \textbf{Model} & \textbf{CSFD} & \textbf{MALL} & \textbf{FCB} \\
    \midrule
    MultiBERT \cite{devlin-etal-2019-bert}          & $81.04\ (\pm0.31)$ & $76.91\ (\pm0.47)$ & $73.24\ (\pm2.37)$ \\
    SlavicBERT \cite{arkhipov2019tuning}            & $82.14\ (\pm0.33)$ & $78.17\ (\pm0.39)$ & $74.19\ (\pm1.85)$ \\
    XLM-RoBERTa-base \cite{conneau2020unsupervised} & $83.74\ (\pm0.39)$ & $77.03\ (\pm0.61)$ & $78.92\ (\pm1.71)$ \\
    Czert \cite{sido2021czert}                      & $83.76\ (\pm0.32)$ & $79.35\ (\pm0.52)$ & $78.34\ (\pm1.59)$ \\
    RobeCzech \cite{straka2021robeczech}            & $85.01\ (\pm0.43)$ & $78.18\ (\pm0.36)$ & $79.12\ (\pm1.39)$ \\
    FERNET-C5                                       & $\mathbf{85.36}\ (\pm0.30)$ & $\mathbf{79.75}\ (\pm0.37)$ & $\mathbf{81.07}\ (\pm1.63)$ \\
    FERNET-News                                  & $84.21\ (\pm0.41)$ & $77.43\ (\pm0.49)$ & $75.94\ (\pm1.85)$ \\
    \midrule
    XLM-RoBERTa-large \cite{conneau2020unsupervised} & $86.03\ (\pm0.32)$ & $79.94\ (\pm0.48)$ & $82.23\ (\pm1.29)$ \\
    \bottomrule
  \end{tabular}
  \caption{Table of results for sentiment analysis task. We report comparison of Transformers in the mean of F1 score on three Czech datasets. Aside from base-sized models, we also report comparison with XLM-RoBERTa-large, which is a much larger model.}
  \label{tab:results_SA}
\end{table}

From the sentiment analysis results (Tab.~\ref{tab:results_SA}), we can see that among base-sized models, our FERNET-C5 model scored best across all datasets. We attribute this superiority to the underlying pre-training data of the model. In all three fine-tuning datasets, we are dealing with text samples written by common people, often under a strong emotions (e.g. angry about a defective product they just bought, delighted about a movie they just saw etc.). It is also a frequent phenomenon (especially in social media domain), that people are writing posts in a common (spoken) Czech, which is extraordinarily different from literary (standard) Czech, used e.g. by journalists. Moreover, diacritical and punctuation marks are often missing in user's posts. Since the C5 dataset was crawled from miscellaneous Czech web pages, all mentioned phenomenons are covered richly in this dataset. Other popular pre-training data sources (Wikipedia, news articles, books etc.) contain mainly standard Czech, which makes models pre-trained from such sources less suitable for this task. Good results scored by multilingual XLM-RoBERTa models, which were trained from Common Crawl data as well, support this hypothesis.

\setlength{\tabcolsep}{8pt}
\begin{table}[th]
  \centering
  \begin{tabular}{ l c c }
    \toprule
    \textbf{Model} & \textbf{CTDC} & \textbf{CN} \\
    \midrule
    MultiBERT \cite{devlin-etal-2019-bert}          & $89.07\ (\pm0.63)$ & $79.83$ \\
    SlavicBERT \cite{arkhipov2019tuning}            & $89.59\ (\pm0.48)$ & $80.46$ \\
    XLM-RoBERTa-base \cite{conneau2020unsupervised} & $88.76\ (\pm0.63)$ & $80.35$ \\
    Czert \cite{sido2021czert}                      & $90.23\ (\pm0.49)$ & $81.50$ \\
    RobeCzech \cite{straka2021robeczech}            & $90.47\ (\pm0.53)$ & $81.20$ \\
    FERNET-C5                                       & $\mathbf{91.25}\ (\pm0.38)$ & $82.13$ \\
    FERNET-News                                  & $90.85\ (\pm0.47)$ & $\mathbf{82.29}$ \\
    \midrule
    XLM-RoBERTa-large \cite{conneau2020unsupervised} & $91.18\ (\pm0.53)$ & $82.78$ \\
    \bottomrule
  \end{tabular}
  \caption{Table of results for multi-label topic identification task. We report comparison of Transformers in the mean of sample-averaged F1 score on two Czech datasets. Aside from base-sized models, we also report comparison with XLM-RoBERTa-large, which is a much larger model.}
  \label{tab:results_MLTI}
\end{table}

From the multi-label topic identification results (Tab.~\ref{tab:results_MLTI}), we can see that both our models scored better than other base-sized models. Since our RoBERTa model was pre-trained from in-domain data for this task (news) and RoBERTa is known to perform better than BERT models, we expected the FERNET-News model to score significantly better than other models. However, we were surprised by the universality of the FERNET-C5 model, which performed comparably well, and on the CTDC dataset, the FERNET-C5 model even outperformed in-domain FERNET-News model. Moreover, FERNET-C5 outperformed even the 5-times larger XML-RoBERTa-large model, however the difference is not statistically significant.

To summarize our results, our models outperformed all public multilingual and monolingual Transformers in the base-model-size category on the text classification tasks. The model pre-trained from 93~GB of filtered Czech Common Crawl dataset (FERNET-C5) demonstrated surprising universality across all domains we were experimenting with.
Results achieved by our base-sized models are also comparable (although slightly worse in most cases) with results scored by 5-times larger model XLM-RoBERTa-large. 

\section{Conclusions}
In this paper, we used large Czech text datasets to pre-train two monolingual Transformers, one from 93~GB of filtered Czech Common Crawl dataset (FERNET-C5) and one from 21~GB of thoroughly cleaned self-crawled news corpus (FERNET-News). The paper describes the dataset preprocessing as well as the models pre-training procedure and evaluates the models on five different Czech text-classification tasks. Our models outperformed all published multilingual and monolingual Transformers in the base-model-size category. The model pre-trained from Common Crawl (FERNET-C5) demonstrated surprising universality across all domains we were experimenting with.

Since Czech is a special language due to its significant differences between spoken and written form, and since internet users often tend to use spoken form with many grammatical errors to write down a content to be classified (social media posts, reviews etc.), it could be very beneficial to cover these phenomena also in the pre-training corpus. Our results showed that model trained from Common Crawl has universal usage across all domains, whereas models pre-trained from popular literary Czech datasets like Wikipedia or news articles, has limited usage for literary datasets only. For example, our FERNET-News pre-trained from thoroughly cleaned news corpus failed to classify correct sentiment of Facebook posts (F1 $75.94$ vs. $81.07$).

Our models are publicly available for research purposes on HuggingFace hub\footnote{\url{https://huggingface.co/fav-kky}}.

\end{document}